\documentclass[letterpaper]{article} 
\usepackage{aaai23}  
\usepackage{times}  
\usepackage{helvet}  
\usepackage{courier}  
\usepackage[hyphens]{url}  
\usepackage{graphicx} 
\usepackage{natbib}  
\usepackage{caption} 
\usepackage{algorithm}
\usepackage{algorithmic}

\usepackage{newfloat}
\usepackage{listings}
\DeclareCaptionStyle{ruled}{labelfont=normalfont,labelsep=colon,strut=off} 
\floatstyle{ruled}
\newfloat{listing}{tb}{lst}{}
\floatname{listing}{Listing}
\title{A Reinforcement Learning Badminton Environment for Simulating Player Tactics (Student Abstract)}
\author {
    Li-Chun Huang,
    Nai-Zen Hseuh,
    Yen-Che Chien,
    \textsuperscript{\rm 1}Wei-Yao Wang,
    Kuang-Da Wang, \\
    Wen-Chih Peng
}
\affiliations{National Yang Ming Chiao Tung University, Hsinchu, Taiwan \\
leo900527@gmail.com, \{august.cs08, dfg15423.cs08, gdwang.cs10\}@nycu.edu.tw, \textsuperscript{\rm 1}sf1638.cs05@nctu.edu.tw, wcpeng@cs.nycu.edu.tw
}

\begin{document}

\maketitle

\begin{abstract}
Recent techniques for analyzing sports precisely has stimulated various approaches to improve player performance and fan engagement.
However, existing approaches are only able to evaluate offline performance since testing in real-time matches requires exhaustive costs and cannot be replicated.
To test in a safe and reproducible simulator, we focus on turn-based sports and introduce a badminton environment by simulating rallies with different angles of view and designing the states, actions, and training procedures.
This benefits not only coaches and players by simulating past matches for tactic investigation, but also researchers from rapidly evaluating their novel algorithms.
Our code is available at https://github.com/wywyWang/CoachAI-Projects/tree/main/Strategic\%20Environment.
\end{abstract}

\section{Introduction}
Nowadays, with the rapid development of analyzing player performance by collecting the past matches they have played, researchers have cooperated with sports teams and players to boost the advancement of sports analytics.
However, it is difficult for novel algorithms to be verified in real-time matches due to the cost and the performance concerns of players.
To mitigate the problem, \citet{DBLP:conf/aaai/KurachRSZBERVMB20} proposed a reinforcement learning football environment, which benefits researchers by reproducing and testing algorithms quickly offline.
Nonetheless, there is no existing environment to develop new ideas in turn-based sports, e.g., badminton, tennis.
Directly using existing environments is not feasible due to the varying nature of different sports.

Therefore, we focus on one of the turn-based sports, badminton, to demonstrate our proposed reinforcement learning environment.
However, there are at least two challenges to describe various factors in a rally.
First, \textbf{3-D Trajectories}: The trajectory of a shuttlecock consists of not only 2-D coordinates but also the height.
The actual height of the shuttlecock cannot be detected precisely due to the regulations in the real-world high-ranking matches and the cost of deploying such advanced techniques (e.g., hawk-eye systems).
Moreover, there are no existing records for the shuttlecock's height, and it is also difficult for domain experts to label the 3-D trajectory, especially with the height.
Second, \textbf{Multi-Agent Turn-Based Environment}: As described in \cite{DBLP:conf/aaai/WangSCP22}, a rally is composed of two players playing alternatively, which is different from the conventional sequence with the same target.
Therefore, it is challenging to design proper states, actions, and rewards for both agents since each agent do the complicated action like returning the shuttlecock and player positioning by taking various observation like the shuttlecock's position and the opponent's into consideration.

To address these issues, we propose a reinforcement learning badminton environment that is equipped with multiple view angles to review a given match (either a simulation or a real match).
In addition, we design the environment based on the multi-agent particle environment (MAPE)  \cite{NIPS2017_68a97503} to describe the process of two agents in a rally.
In this manner, our badminton environment is able to not only support coaches and players to review and investigate tactics of players in a more flexible way, but also provides researchers with an interface place to quickly demonstrate new algorithms.
For a more detailed illustration, please refer to our demonstration here\footnote{https://youtu.be/WRPcbalb6yc.}.

\section{Approach}

\subsection{Dataset Collection}
We use the dataset collected by previous research \cite{DBLP:conf/aaai/WangSCP22}, which includes 75 high-ranking matches from 2018 to 2021 by 31 players from men's singles and women's singles games labeled by domain experts with the BLSR format \cite{DBLP:conf/icdm/WangCYWFP21,10.1145/3551391}.
The dataset includes the positions of the players and the shuttlecock as 2-D coordinates, the timestamp of each ball round, the type of ball, the scores, and the motions of players.
We aimed to learn the tactics from different players by these datasets.

\subsubsection{Mimicking Actual Ball Height}
We lack information about the shuttlecock's height in the collected dataset (we only know a label describing whether the hit point is above the net or not).
To simulate the actual height, we set the height of each shot type with the average height below the net and standard deviation, and then use normal distribution as the corresponding distribution.

\subsection{Reinforcement Learning Badminton Environment}
As the tactics vary according to individual player, we have to design a process in a rally that is able to mimic players while considering different factors.
Specifically, our environment is based on MAPE, which supports multi-agent training.

\subsubsection{Environment Design}
The environment is designed following a regular real-world badminton court, which includes two players from each side, a shuttlecock, the net and the boundary.
To have a better visualization experience and adapt different application scenarios, we proposed multi-view observation options, which enables the user to monitor the playing process (training process when training agents) through the side view or the top view.
To cope with this limitation, we designed a size shrinking method to illustrate the height of the shuttlecock.
Specifically, the rendering object is bigger if the shuttlecock is closer to the player, and smaller otherwise.

\subsubsection{Turn-Based Procedure}
As badminton is a fast-paced sport, it is difficult for the agents to move instantaneously.
Therefore, our goal is to make the agent focus on learning the tactics of the badminton player instead of playing badminton.
We therefore simplify the real-time game into a turn-based environment.
The procedure in a rally is as follows: 1) Assume that the shuttlecock is served by player A. In this sub-step, player A, as an agent, will decide the landing position of the shuttlecock, the ball type to hit, and the defense position to go to after returning the ball.
On the other hand, player B, as an opponent agent, will decide the target position to go to in order to return the shuttlecock.
2) The environment will simulate the player's move and the trajectory of the shuttlecock until the shuttlecock reaches the defense region of the opponent.
3) At the moment the shuttlecock enters the opponent's defense region, the simulation will stop.
On the other hand, the player will also decide the target position to go to in order to return the shuttlecock.
4) After receiving the players' decision, the environment will keep simulating until the shuttlecock falls into the proper region, that is close enough to the opponent and the height of the shuttlecock is reasonable for the type of shot the opponent is returning.
5) The step is finished, so the roles of the players swap. The environment executes the returning action and goes back to Step 2 until the rally is finished.

\subsubsection{Simulation}
To produce a realistic environment and enhance the reference value of the environment in reality, we apply the meta-parameter based on the match dataset.
The meta-parameter we tuned based on the dataset includes the player speed, the defense range of the players, the returning region distribution of different ball types, and other physical parameters of the shuttlecock.
Furthermore, we follow \cite{Chen2009ASO} to simulate the shuttlecock trajectory.

\section{Preliminary Results}

\begin{figure}
    \centering
    \includegraphics[height=!, width=\linewidth, keepaspectratio]{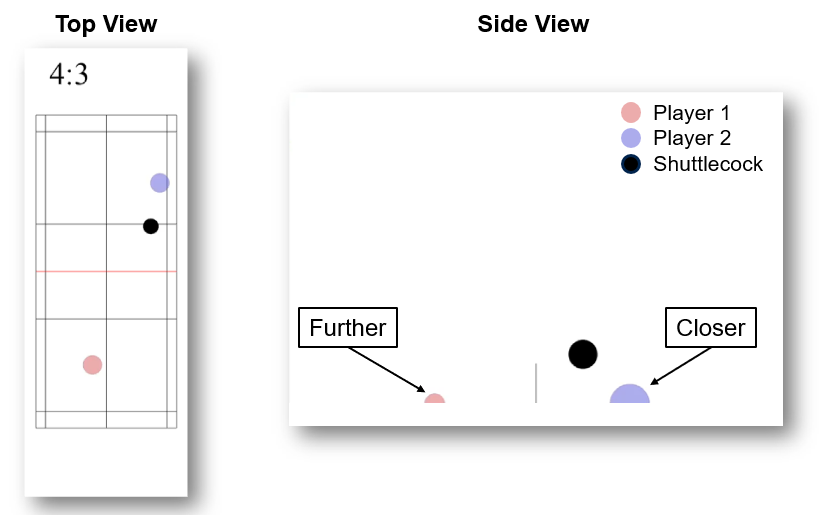}
    \caption{The schematic of the reinforcement learning badminton environment with two supporting views.}
    \label{fig1: multi-view}
\end{figure}


\noindent\textbf{Multiple Angles of View. }
Figure \ref{fig1: multi-view} illustrates our proposed badminton environment equipped with different views, which enables researchers and domain experts to observe the playing procedure.
\noindent\textbf{Multi Agent. }
In general reinforcement learning (RL) environments, there is just one agent to interact with an environment in a match.
However, badminton games are usually for two or four players to play, so we built our environment based on MAPE to achieve this function.
Our environment is able to train not just one, but two or three or four agents in the same match, and can deal situations like a training agent versus with an expert player or two agents controlling two players on the same side respectively in doubles games.
\noindent\textbf{Recording Match Data. }
One of the characteristics in our environment is that it records the match data through the matches.
This technique benefits researchers with not only data augmentation but also debugging for improving training policies.

\bibliography{aaai23}

\end{document}